%% file: main.tex
\documentclass[11pt]{article}

\usepackage{acl}

\usepackage{svg}
\usepackage{times}
\usepackage{latexsym}
\usepackage{amssymb}

\usepackage{subcaption}

\usepackage[T1]{fontenc}

\usepackage[utf8]{inputenc}

\usepackage{microtype}

\usepackage{inconsolata}
\usepackage{algorithm}
\usepackage{algorithmicx}
\usepackage{algpseudocode}
\usepackage{enumitem}

\usepackage{graphicx}

\title{An \textit{In-Vitro} Study on Cross-Lingual Generalization in Language Models}

\author{%
  Adrian Cosma \\
  Dalle Molle Institute for Artificial Intelligence (IDSIA) \\
  \texttt{adrian.cosma@idsia.ch} \\
}

\usepackage[most]{tcolorbox}

\definecolor{takeawaycolor}{HTML}{8ABEE5}
\tcbset {
  base/.style={
    title={#1},
    colbacktitle=takeawaycolor,
  }
}
\newtcolorbox{mainbox}[1]{
  base={#1}
}

\newcounter{takeaway}

\definecolor{BG}{HTML}{F8F8F8}
\definecolor{Acolor}{HTML}{287AB8}
\definecolor{Bcolor}{HTML}{DA4167}
\usepackage{longtable}
\usepackage{booktabs}
\usepackage{array}

\newcommand{\LA}{{\color{Acolor}$\mathbf{A}$}}
\newcommand{\LB}{{\color{Bcolor}$\mathbf{B}$}}
\newcommand{\LBm}{{\color{Bcolor}$\mathbf{B}^\dagger$}}

\begin{document}
\maketitle



\begin{abstract}
Cross-lingual transfer in language models is difficult to study in natural corpora because lexical overlap, morphology, data imbalance, and tokenization are entangled. We introduce an \textit{in-vitro} framework with two procedurally generated languages that share the same ontology, typed grammar, and compositional structure, but differ in surface realization. This lets us independently vary lexical distance, minority-language proportion, tokenizer training regime, and vocabulary size, while evaluating transfer on a masked minority-language condition whose lexical forms are never observed during training. Across 700 controlled runs, we find that transfer is governed less by tokenizer balance or raw lexical similarity than by whether tokenization preserves reusable cross-lingual substructure. Smaller vocabularies often improve masked transfer by keeping words decomposable into shared fragments, whereas larger vocabularies can turn forms into language-specific atoms. We further show that transfer emerges as a staged process: grammatical and type-level competence precede masked lexical generalization. Finally, we attempt to explain this mechanism through \textit{tokenizer bridges} and show that bridge strength correlates strongly with masked reachability.
\end{abstract}

\section{Introduction}
\label{sec:intro}
\input{sections/1.intro}

\section{Related Work}
\label{sec:related}

\input{sections/2.relatedwork}

\begin{figure*}[hbt!]
    \centering
    \includesvg[width=1.0\linewidth]{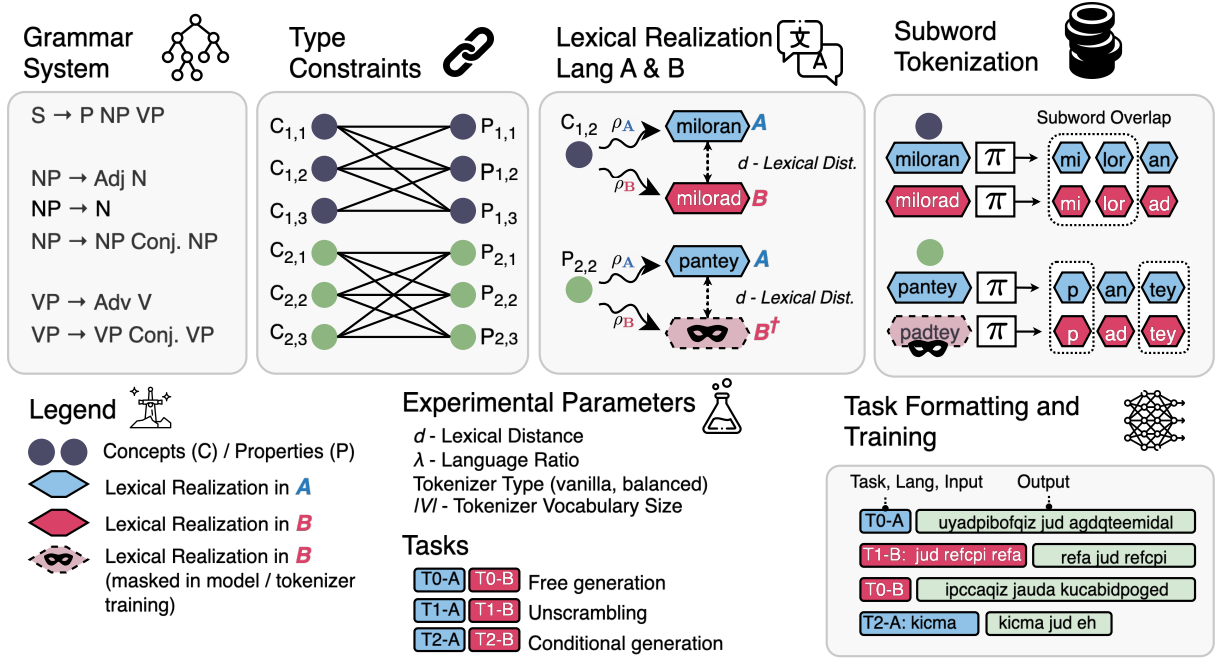}
    \caption{\textbf{Overview of the in-vitro cross-lingual setup.} Two procedurally generated languages, \LA{} and \LB{}, share the same underlying grammar, type constraints, and concept–property ontology, but differ in their lexical realizations. Cross-lingual difficulty is controlled through lexical distance $d$, minority-language proportion $\lambda$, tokenizer type, and tokenizer vocabulary size. A subset of minority-language lexical forms is withheld to define the masked condition \LBm, enabling evaluation of whether models transfer shared compositional structure rather than memorizing surface forms. Subword tokenization mediates overlap between the two languages.}
    \label{fig:diagram}
\end{figure*}

\section{In-Vitro Cross-Lingual Setup}
\label{sec:method}
\input{sections/3.method}

\section{Results}
\label{sec:exps}
\input{sections/4.results}

\subsection{Token bridges facilitate cross-lingual transfer}
\label{sec:theory}

\input{sections/3.1.theory}

\section{Conclusions}
\label{sec:conc}
\input{sections/5.conclusions}

\section*{Limitations}
\label{sec:lim}
\input{sections/6.limitations}

\bibliography{refs}

\appendix
\section{Appendix}
\label{sec:appendix}
\input{sections/7.appendix}

\end{document}

%% file: sections/1.intro.tex
Language models (LMs) exhibit cross-lingual transfer: training on one language can improve performance in another even without direct supervision in the target language \cite{zhao2024how,de2024measuring}. Yet the mechanism behind this transfer remains unclear.  In this work, we ask the question: \textit{How are semantic language capabilities transferred during training through surface forms?} 

Prior work suggests that multilingual models develop both language-specific and shared language-agnostic representations \cite{zhao2024how,de2024measuring}, but existing evidence remains mixed on the role of language similarity and data composition \cite{malkin2022balanced,limisiewicz2023you}, while tokenization has repeatedly emerged as a critical interface between surface form and meaning \cite{rust2021good,patil2022overlap,hammerl2025beyond}. However, in natural multilingual corpora, lexical overlap, morphology, script, domain, corpus size, and tokenization all vary at once, making it difficult to determine which factors actually enable knowledge learned in one language to become useful in another \cite{blevins2022language}.

In this work, we study cross-lingual generalization \textit{in vitro}. We build a framework to procedurally generate two languages, \LA{} and \LB{}, that share the same underlying ontology, typed grammar, and compositional structure, but differ in their lexical realization. This lets us parameterize the generation process with three factors that are usually confounded in natural data: lexical distance between languages, the relative amount of training data in each language, and the granularity induced by subword tokenization. We further define a masked minority-language condition, \LBm{}, by withholding a subset of lexical realizations from language \LB{} during training. Success on \LBm therefore requires the model to transfer shared compositional structure across languages rather than merely memorize minority-language surface forms. Our setup enables the study of multilingual transfer in a controlled manner, pertaining to the larger context of \textit{learning mechanics} \cite{michaud2023quantization,liu2025physics,allen2023physics}.

In particular, our setup enables us to test the following hypotheses from literature:
\begin{enumerate}
    \item[$\mathcal{H}_1$] \textbf{\textit{Tokenizer Influence:}} Cross-lingual transfer depends less on tokenizer balance itself than on whether tokenization preserves shared subword bridges between languages~\cite{rust2021good,patil2022overlap,hammerl2025beyond}
    \item[$\mathcal{H}_2$] \textbf{\textit{Lexical Similarity:}} Lexically similar languages will result in easier cross-lingual transfer~\cite{kunchukuttan2018leveraging,Gooskens_Swarte_2017,hernandez2021scaling}
    \item[$\mathcal{H}_3$] \textbf{\textit{Language Proportions:}} Increasing the amount of minority-language data should improve target language capabilities, though evidence suggests that the benefits are condition-dependent ~\cite{malkin2022balanced,limisiewicz2023you}.
    \item[$\mathcal{H}_4$] \textbf{\textit{Skill Cascade:}} Cross-lingual transfer is preceded by a cascade of skills that are required: learning the majority language (grammar, type constraints), learning the minority language (grammar, type constraints), then generalize to unseen concepts~\cite{blevins2022analyzing,lubana2024percolation,wang2024probing}.
\end{enumerate}

We attempt to formalize the cross-lingual transfer mechanism through \textit{tokenizer bridges}. For each masked lexical item, we measure how strongly its tokenization overlaps with observed lexical forms under the same tokenizer. We find that this bridge strength correlates with masked reachability, suggesting that transfer improves when unseen target-language forms remain connected to observed forms through shared subword fragments. The resulting view is that multilingual generalization is a property of the tokenized topology through which languages are presented to the model.

Our contributions are as follows:
\begin{enumerate}
    \item We introduce a controlled \textit{in-vitro} framework for studying cross-lingual transfer under independently varied lexical distance, language proportion, tokenizer regime, and vocabulary size.
    \item We define a masked minority-language condition that isolates cross-lingual generalization from memorization of observed target-language forms.
    \item We show that smaller vocabularies and reusable subword structure can improve masked transfer, while larger vocabularies may suppress it by isolating languages into more language-specific units.
    \item We propose \textit{tokenizer bridges} as a potential explanation for masked cross-lingual transfer and show that bridge strength predicts masked reachability.
\end{enumerate}

%% file: sections/2.relatedwork.tex
Prior work suggests that multilingual LMs develop both language-specific and language-agnostic representations \cite{zhao2024how,de2024measuring}. Evidence for shared internal structure comes from probing studies identifying overlapping or language-specific neurons \cite{wendler-etal-2024-llamas,zhao2024how}, from synthetic pretraining experiments showing transfer across unrelated surface forms \cite{papadimitriou2020learning}, and from studies of multilingual pretraining mixtures \cite{de2024measuring}. These findings suggest that LMs can abstract away from surface realization and capture latent structure shared across languages.

At the same time, cross-lingual transfer is strongly shaped by the tokenization interface through which text enters the model. Unlike humans, whose mutual intelligibility is mediated by vocabulary, phonology, and syntax \cite{gooskens2021mutual,escudero2004bridging,Gooskens_Swarte_2017}, LMs operate over subword units induced by data-driven tokenizers. This can introduce systematic unfairness across languages, especially when tokenizer training data is dominated by high-resource languages \cite{limisiewicz2023tokenization,unfairness-tokenizer,arnett2025explaining,hammerl2025beyond,patil2022overlap}. Prior work has therefore explored both tokenizer-level interventions \cite{abagyan2025one,jabbar2023morphpiece} and architecture-level alternatives \cite{xue2022byt5,ahia2024magnet,cosma2025strawberry}. However, the interaction between language similarity, data proportions, and tokenization remains difficult to isolate in natural multilingual corpora.

\noindent \textbf{This work.} We study multilingual transfer in a controlled synthetic setting that isolates language similarity, training proportions, and tokenizer parameters. Instead of relying on natural languages, where domain, ontology, and lexical similarity are heavily confounded, we construct programmatic languages over a shared semantic space \cite{taguchi2026creatingconlangsprobemetalinguistic}. Building on \citet{lubana2024percolation}, we generate controlled lexical realizations on top of a Probabilistic Context-Sensitive Grammar with type constraints. Our evaluation tests generalization to concept realizations unseen during training, across multiple abstraction levels: word validity, grammaticality, constraint satisfaction, and lexical reachability of masked concepts. This allows us to examine when multilingual transfer emerges during training, and how semantic structure is transferred across distinct lexical forms. More broadly, our setup follows a line of work using formal grammars and synthetic data to expose learning dynamics that are difficult to identify in natural data with many confounders \cite{papadimitriou2023injecting,lubana2024percolation,allen2023physics,hu2025between,blevins2022language}.

%% file: sections/3.method.tex
\subsection{Setup Formalization}
\label{sec:setup-formalization}

Our objective is to isolate how cross-lingual transfer depends on lexical similarity between languages, language proportions in the training corpus, and subword tokenization granularity. 

In our framework, we model language as a structured symbolic system specified by \textit{(i)} a vocabulary of abstract symbols, \textit{(ii)} a type signature over those symbols, \textit{(iii)} a grammar that defines syntactically well-formed sentences, and \textit{(iv)} a realization function that maps abstract symbols to surface language-specific word forms. We extend the setup of \citet{lubana2024percolation} to account for multi-language realizations on top of the typed grammar. Readers are referred to the original work for more details on the framework implementation. As such, we do not treat language in the full sociolinguistic sense~\cite{Ballard_1980}, but rather as a finite generative system with a shared abstract layer and language-specific realizations. 

We study cross-lingual generalization in a controlled multilingual pretraining setting with two procedurally generated languages, denoted with \LA~and \LB. For notational convenience, we index them by $\ell \in \{$\LA, \LB$\}$. We also define a masked evaluation condition, denoted \LBm, which is derived from \LB~ by withholding a subset of minority-language lexical realizations during training. All three conditions share the same underlying ontology and compositional rules, but differ in their lexical surface realization. We show in Figure \ref{fig:diagram} a diagram of our setup.

\paragraph{Symbolic system and language realization.}

A language is a 6-tuple: $\ell = (\mathcal{V}, \mathcal{T}, \tau, G, \Gamma, \rho_{\ell})$, where:

\begin{itemize}
    \item $\mathcal{V}$ is a finite inventory of abstract lexical symbols;
    \item $\mathcal{T}$ is a finite set of types;
    \item $\tau : \mathcal{V} \to \mathcal{T}$ assigns each symbol a type;
    \item $G$ is a probabilistic grammar over $\mathcal{V}$ that generates well-formed symbolic sequences;
    \item $\Gamma \subseteq \mathcal{C} \times \mathcal{P}$ is an ontology of semantically valid concept--property pairs, where $\mathcal{C}, \mathcal{P} \subseteq \mathcal{V}$.
    \item $\rho_\ell : \mathcal{V} \to \Sigma_\ell^*$, a realization function which maps abstract symbols to surface forms in language $\ell$ to a vocabulary $\Sigma_\ell$.
\end{itemize}

In particular, we define $G$ as a \emph{Probabilistic Context-Sensitive Grammar} (PCSG) (see Appendix \ref{sec:appendix}). For the ontology, let $\mathcal{C} = \{c_1,\dots,c_{N_C}\}$ and $\mathcal{P} = \{p_1,\dots,p_{N_P}\}$ be finite sets of \emph{concepts} and \emph{properties}, respectively. We assume a typed relational structure $\Gamma \subseteq \mathcal{C} \times \mathcal{P}$, where $(c,p)\in\Gamma$ indicates that property $p$ is semantically valid for concept $c$. This defines the ontology shared by all language conditions. Intuitively, $\Gamma$ specifies which attributes may describe which entities. For example, this encodes relationships such as \textit{(hiker, climbs)} as valid but \textit{(table, eats)} as invalid.

This separates syntax from semantics and from lexical realization: $G$ determines which symbolic sequences are structurally well formed, while $\Gamma$ constrains which concept-property combinations are semantically valid.

\paragraph{Symbolic sentence generation.}
A sentence is generated in three stages. First, a symbolic template $z = (v_1,\dots,v_n)\in \mathcal{V}^n$ is sampled from the grammar $G$. 

Second, the sampled symbols must satisfy the relevant type and ontology constraints. For example, if $z$ describes a concept $c\in\mathcal{C}$ with a property $p\in\mathcal{P}$, then $(c,p)\in\Gamma$ must hold. Third, the symbolic sentence is realized in language $L$ by applying $\rho_{\ell}$ tokenwise: $\rho_{\ell}(z) = \bigl(\rho_{\ell}(v_1), \dots, \rho_{\ell}(v_n)\bigr)$.

This yields a corpora in different languages that are semantically aligned at the symbolic level but may differ arbitrarily in surface form.

\subsection{Multilingual Dataset Generation}
\label{sec:dataset-generation}

\paragraph{Lexical realization and lexical distance.}

The two languages \LA{} and \LB{} share the same abstract vocabulary $\mathcal{V}$, type system $\tau$, grammar $G$, and ontology $\Gamma$, but differ in their realization functions $\rho_{\color{Acolor}\mathbf{A}}$ and $\rho_{\color{Bcolor}\mathbf{B}}$. We construct lexical realization by assuming languages are agglutinative \cite{agglutinative} and constructed from concatenation of different morphemes: language-specific prefixes, stems and suffixes. Furthermore, each language has its specific pre-defined frequency distribution of characters.

Experimentally, we control cross-lingual lexical similarity through a scalar parameter $d \in [0,1]$, where larger values of $d$ correspond to stronger perturbations of the latent stems and therefore greater lexical divergence between \LA{} and \LB{}. This roughly corresponds to a lexical similarity distance \cite{levenshtein1966binary} used in linguistic studies \cite{dinu2023robocop}. For example, we could have the stem \textit{"caqujna"}, which leads to the two words \textit{"naw\underline{caqjna}"} and \textit{"ag\underline{caqujna}"} in \LA{} and \LB{}, respectively. The stem is being modified and each language added its own prefix. In Appendix \ref{sec:appendix} we provide the pseudocode for word formation and show several examples of generated sentences. We discuss limitations of this approach in the \nameref{sec:lim} section.

\paragraph{Training corpus mixture.}
Let $\mathcal{D}_A$ and $\mathcal{D}_B$ denote corpora of realized training sequences in \LA{} and \LB{}, respectively, both generated from the same underlying symbolic process. We vary the language mixture by controlling the corpus ratio $
\frac{|\mathcal{D}_B|}{|\mathcal{D}_A|} = \lambda$. We maintain \LB{} as the minority language, by ensuring $\lambda \in (0, 0.5]$. Varying $\lambda$ allows us to separate effects of lexical similarity from effects of training language proportions.

\paragraph{Masked minority-language condition.}
To test genuine cross-lingual transfer rather than direct memorization of minority-language surface forms, we define a masked evaluation condition \LBm{}. Let $\mathcal{V}^{\dagger} \subset \mathcal{V}$ be a subset of lexical symbols whose realizations in language \LB{} are withheld during model training. We then define

$$\mathcal{D}_{{\color{Bcolor} \mathbf{B}^\dagger}}
= \left\{
x \in \mathcal{D}_{{\color{Bcolor} \mathbf{B}}}:
x\text{ has } \rho_{{\color{Bcolor} \mathbf{B}}}(v) \text{ for some } v \in \mathcal{V}^{\dagger}
\right\}$$

The remaining minority-language examples are

$\mathcal{D}_{\color{Bcolor}\mathbf{B}}^{\mathrm{seen}} = \mathcal{D}_{\color{Bcolor}\mathbf{B}} \setminus \mathcal{D}_{{\color{Bcolor}\mathbf{B}^\dagger}}$, and the training corpus is $\mathcal{D}_{\mathrm{train}} = \mathcal{D}_{\color{Acolor}\mathbf{A}} \cup \mathcal{D}_{\color{Bcolor}\mathbf{B}}^{\mathrm{seen}}.$

Crucially, the underlying symbolic structures associated with $\mathcal{V}^{\dagger}$ may still be observed through \LA{}. Success on \LBm therefore requires transferring shared compositional and semantic structure, rather than relying on direct exposure to the withheld \LB{} lexical forms. We mask 25\% of the property symbols for \LB{}. 

\paragraph{Tokenizer.}
Before model training, realized text is segmented by a subword tokenizer
$\pi : \Sigma^*_{\color{Acolor}\mathbf{A}} \bigcup \Sigma^*_{\color{Bcolor}\mathbf{B}} \to V^*$, where $V$ is a learned BPE vocabulary of size $|V|$. We consider two tokenizer-training regimes: \emph{vanilla} ($1:\lambda$) and \emph{balanced} ($1:1$). In the more realistic \emph{vanilla} condition, tokenizer training uses the same language mixture as model training; in the \emph{balanced} condition, tokenizer training uses equal proportions of examples from both languages and is decoupled from the model-training mixture. In particular, in both cases, we train a BPE tokenizer \cite{Gage1994ANA} with whitespace pre-tokenization on a dataset of fixed size. Since tokenization determines the atomic units through which the model observes surface form, it acts as a bottleneck between lexical realization and cross-lingual abstraction \cite{limisiewicz2023tokenization,unfairness-tokenizer,patil2022overlap}. In contrast to other works \cite{de2024measuring} that use a byte tokenizer \cite{xue2022byt5}, we opt to vary tokenization parameters to control for the effect of subword tokenization. 

\paragraph{Model Training.}
We train a small decoder-only transformer LM on sentences generated using the procedure described above. Following \citet{lubana2024percolation}, we construct several tasks for model training and evaluation: free generation (T$_0$), unscrambling (T$_1$) and conditional generation (T$_2$). For each task-language pair, we have special tokens to condition the model on which task it is supposed to perform and in which language. As such, we have a total of 6 such special tokens (3 tasks $\times$ 2 languages). Examples are shown in Figure \ref{fig:diagram}.

\paragraph{Problem statement.}
Given two languages with a shared symbolic system, a lexical-distance parameter $d$, a corpus-mixture parameter $\lambda$, and a tokenizer $\pi$, we ask: \textit{under what conditions does knowledge acquired from the majority language \LA{} become available in the minority language \LB{}, especially on the masked condition \LBm{}?} Because $\mathcal{D}_{{\color{Bcolor}\mathbf{B}^\dagger}}$ is excluded from training, performance on \LBm{} is a direct measure of cross-lingual generalization through shared structure.

\subsection{Evaluation Setup}
Instead of simply looking at perplexity, which is an insufficient measure of downstream performance~\cite{velivckovic2026perplexity}, we primarily measure grammaticality and type constraints satisfaction~\cite{lubana2024percolation}. Our evaluation is split into \textit{intra-language} performance, looking individually at how performance on \LA{} and \LB{} progresses, and \textit{cross-language} generalization by measuring performance on the masked lexical realizations in \LBm{}. We measure performance for task $T_1$ (unscrambling) for both maximal metric value and emergence step \cite{wei2022emergent,berti2025emergent}. We consider the definition of emergence by \citet{anderson1972more}\footnote{"Emergence is when quantitative changes in a system result in qualitative changes in behavior." \cite{anderson1972more}}, and we compute emergence step as the first training step for which a particular metric achieves a value greater than 0.02. 

Furthermore, we introduce a modified version of \textit{concept reachability} \cite{lubana2024percolation} that is designed to accommodate tokenized concepts. In particular, we measure \textbf{Top-$K$ Reachability.} The model is conditioned with the templates "\texttt{$T_0$-\LA{} <subject\_id> is \rule{0.5cm}{0.4pt}}" or "\texttt{$T_0$-\LB{} <subject\_id> is \rule{0.5cm}{0.4pt}}" and is required to generate the tokens to a valid descriptive property to the specified \texttt{subject\_id} in the target language. Since such properties could be lexically realized in multiple tokens, we test whether there exists a path where every next token is in the top-$K$ next tokens under the evolving context. We fix $K = 0.1 \times |V|$, to compare between tokenizer vocabulary sizes. Subjects are chosen such that the realized versions of the associated properties are not seen during model training for \LB, but are seen in \LA. This directly measures cross-lingual semantic transfer through lexical realizations. We provide pseudocode in Appendix \ref{sec:appendix}.

%% file: sections/4.results.tex
\begin{figure*}[hbt!]
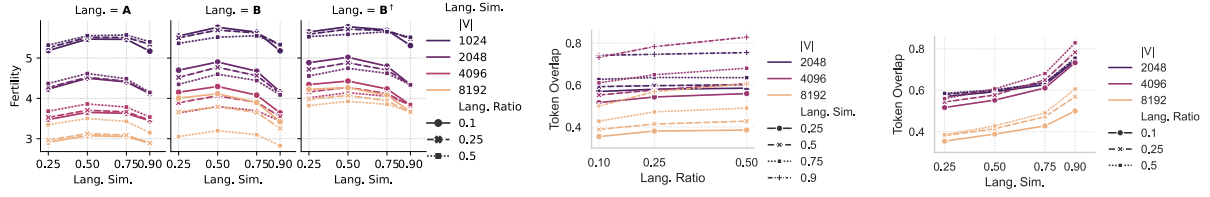

    \centering
    \begin{subfigure}{0.44\linewidth}
        \centering
        \includesvg[width=1.0\linewidth]{figures/tokenizer/fragmentation.svg}
        \caption{Tokenizer fertility for \LA{}, \LB{}, and \LBm{} based on lexical similarity, language ratio, and vocabulary size.}
        \label{fig:fragmentation-main}
    \end{subfigure}
    \begin{subfigure}{0.27\linewidth}
        \centering
        \includesvg[width=\linewidth]{figures/tokenizer/token-vocab-overlap-by-ratio.svg}
        \caption{Vocabulary overlap increases with amount of \LB{} data.}
        \label{fig:vocab-overlap-ratio}
    \end{subfigure}
    \hfill
    \begin{subfigure}{0.27\linewidth}
        \centering
        \includesvg[width=\linewidth]{figures/tokenizer/token-vocab-overlap-by-similarity.svg}
        \caption{Vocabulary overlap varies with lexical similarity.}
        \label{fig:vocab-overlap-similarity}
    \end{subfigure}
    \caption{Tokenizer fertility and vocabulary overlap under different multilingual conditions.}
    \label{fig:fragmentation}
\end{figure*}

\begin{figure*}[hbt!]
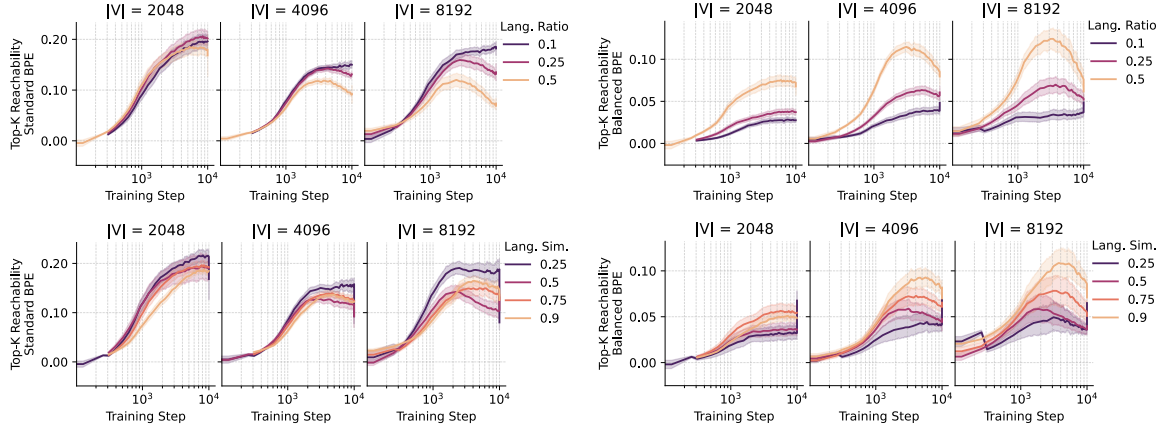

    \centering
    \includesvg[width=0.48\linewidth]{figures/teaser-reachability-vanilla-bpe.svg}
    \includesvg[width=0.48\linewidth]{figures/teaser-reachability-balanced-bpe.svg}
    \includesvg[width=0.48\linewidth]{figures/teaser-reachability-vanilla-bpe-sim.svg}
    \includesvg[width=0.48\linewidth]{figures/teaser-reachability-balanced-bpe-sim.svg}
    \caption{\textbf{Masked top-$K$ reachability across training.} We report Top-$K$ reachability in the masked minority-language condition \LBm{} across tokenizer vocabulary sizes, language proportions, lexical similarities, and tokenizer training regimes. Smaller vocabularies generally yield higher masked reachability, while larger vocabularies suppress transfer by inducing more language-specific units. Balanced tokenization improves reachability in some low-resource settings, but the effect is not uniform across vocabulary sizes.}
    \label{fig:reachability_main}
\end{figure*}

\begin{figure*}[hbt!]
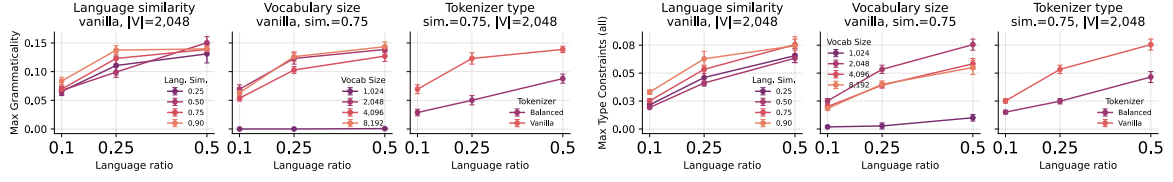
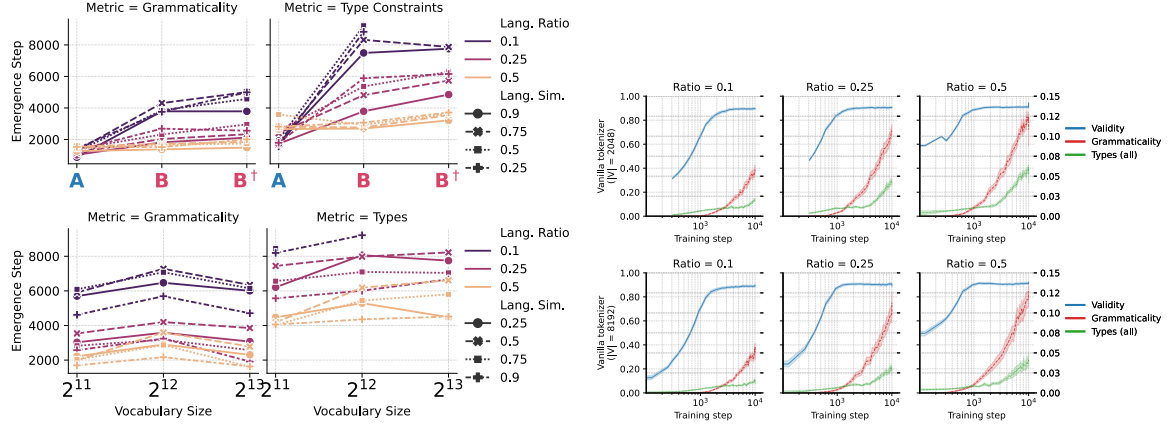

    \centering
    \begin{subfigure}{\linewidth}
        \centering
        \includesvg[width=0.48\linewidth]{figures/results/max-perf-max-grammaticality-vanilla-2048-0.25.svg}
        \includesvg[width=0.48\linewidth]{figures/results/max-perf-max-type-constraints-all-vanilla-2048-0.25.svg}
        \caption{Grammaticality and type-constraint satisfaction across proportions, lexical similarity, vocabulary size, and tokenizer type.}
        \label{fig:max-performance}
    \end{subfigure}
    \begin{subfigure}{0.48\linewidth}
        \centering
        \includesvg[width=1.0\linewidth]{figures/results/language-capabilities-vanilla-2048.svg}
           \includesvg[width=1.0\linewidth]{figures/results/emergence-average.svg}
        \caption{Average emergence time for grammaticality and type-constraint satisfaction across languages for vanilla tokenizer and $|V|= 2048$ (top) and across vocabulary sizes for \LBm{} (bottom). Lower values indicate earlier emergence during training.}
        \label{fig:emergence-average}
    \end{subfigure}
    \begin{subfigure}{0.48\linewidth}
        \includesvg[width=1.0\linewidth]{figures/results/condensed-grammar-B-masked-vanilla-vocab2048.svg}
        \includesvg[width=1.0\linewidth]{figures/results/condensed-grammar-B-masked-vanilla-vocab8192.svg}
    \caption{Training dynamics for validity, grammaticality, and type-constraint satisfaction in the masked condition \LBm{}. Capabilities emerge in a cascaded manner, before type constraints can be satisfied, the model learns to construct valid words and grammatically correct sentences.}
        \label{fig:grammar}
    \end{subfigure}
    \caption{\textbf{Structural capabilities emerge before masked lexical transfer.} The results support a staged view of cross-lingual transfer: models first acquire shared structural constraints before generalizing to unseen minority-language lexical realizations.}
    \label{fig:results}
\end{figure*}

We study cross-lingual generalization along three axes: tokenizer training regime, tokenizer vocabulary size, and the relation between language proportion and lexical similarity. We first analyse the tokenizer-induced structure of the two languages, then evaluate whether this structure supports transfer to the masked minority-language condition \LBm{}. We then compare the emergence of different capabilities to determine whether masked lexical generalization appears as an isolated effect or as the last stage of a broader learning trajectory. Finally, we attempt to explain masked cross-lingual transfer through \textit{tokenizer bridges}.

\subsection{Main results}
\paragraph{Tokenization drives cross-lingual structure ($\mathcal{H}_1$).} In Fig.~\ref{fig:fragmentation} we show fertility \cite{rust2021good} and vocabulary overlap between the two languages, across multiple language parameters (see Appendix \ref{sec:appendix}, Fig. \ref{fig:fragmentation-cont} for similar plots with continuation rate as the primary metric). The tokenizer does not expose the model to the two languages symmetrically. The minority language is more fragmented than the majority language. This effect is strongest at smaller language ratios, where fewer \LB{} examples are available during tokenizer training. The overlap plots in Fig.~\ref{fig:fragmentation} show that increasing the tokenizer vocabulary size generally reduces shared subword structure between the two languages. As the vocabulary grows, BPE increasingly memorizes longer language-specific units rather than preserving reusable pieces shared across \LA{} and \LB{}. This creates a tension: larger vocabularies reduce fragmentation within each language, but also make the two languages less coupled at the subword level. Since transfer to \LBm{} requires the model to use information acquired from \LA{}, this loss of shared substructure can harm cross-lingual generalization.

\paragraph{Balanced tokenizer training is not uniformly better ($\mathcal{H}_1$).} In Fig.~\ref{fig:reachability_main} we show a comparison between masked concept reachability across language proportions, tokenizer vocabulary sizes, and tokenizer types. In most settings, vanilla tokenization matches or exceeds the balanced tokenizer despite being trained on a skewed mixture. This indicates that the relevant factor is whether the resulting segmentation preserves subword units that can be reused across languages. A balanced tokenizer can still fail if it allocates capacity to language-specific whole-word or near-whole-word tokens, while a vanilla tokenizer can remain useful if its skewed training distribution induces smaller shared fragments.

Equal tokenizer exposure to both languages is not sufficient for transfer. What matters is the structure of the induced vocabulary, especially whether the segmentation creates bridges between \LA{} and unseen lexical forms in \LBm{}. In Figure \ref{fig:max-performance}, we show that maximum performance for grammaticality and type constraints satisfaction is consistently lower for the balanced tokenizer.

\paragraph{Larger vocabularies suppress masked transfer. ($\mathcal{H}_1$)} Across tokenizer regimes, increasing vocabulary size tends to reduce masked concept reachability in \LBm{} (Fig.~\ref{fig:reachability_main}). This pattern is clearest in the reachability curves, where smaller vocabularies produce higher top-$K$ reachability than larger vocabularies under otherwise comparable conditions. In our setting, the transfer problem depends on whether the model can connect unseen \LB{} lexical realizations to structure learned from \LA{}. When the vocabulary is large, BPE tends to represent words as language-specific atoms. This reduces the number of shared intermediate units through which the model can generalize from one language to the other. At the extreme, a word-level vocabulary would eliminate almost all useful overlap between unrelated surface forms, leaving no subword path for transfer.

Thus, vocabulary size has opposite effects on two objectives: it can make individual-language modeling more efficient while making cross-lingual abstraction harder. For masked lexical transfer, the useful vocabulary is not necessarily the largest one, but the one that preserves reusable sublexical structure.

\paragraph{Language proportion controls transfer more strongly than lexical similarity ($\mathcal{H}_2$, $\mathcal{H}_3$).} The language-mixture parameter $\lambda$ has a consistent effect on \LBm{} performance. Increasing the amount of minority-language data improves grammaticality, type-constraint satisfaction, and masked reachability (Figure \ref{fig:max-performance}). This is expected: more \LB{} data gives the model more evidence about the target language's surface regularities, even though the specific masked lexical forms are withheld. In .

By contrast, lexical similarity has a weaker and less consistent effect. Although higher similarity sometimes improves reachability, the differences across similarity values are smaller than the differences induced by tokenizer vocabulary size and language proportion. This suggests that surface similarity alone is not the dominant driver of transfer. The model appears to rely more on abstract structure and tokenizer-mediated overlap than on raw character-level proximity between the two realization functions.

Similarity aids the creation of shared subword structure, but it is not by itself sufficient. Two languages may be character-level similar yet poorly aligned under BPE, or less similar yet still connected through reusable fragments. In this setting, tokenization is the mechanism through which lexical similarity becomes available to the model.

\paragraph{Transfer emerges as a staged process ($\mathcal{H}_4$).} In Fig.~\ref{fig:emergence-average} and \ref{fig:grammar} we show that cross-lingual generalization emerges in a cascade. The model first learns high-level well-formedness in the majority language \LA{}, it then improves on the observed minority language \LB{}, and only later shows meaningful generalization to the masked condition \LBm{}. The same ordering appears across both grammaticality and type-constraint satisfaction. Grammaticality and type satisfaction rise earlier \cite{lubana2024percolation}, while masked reachability remains lower and more sensitive to tokenizer design. 

The results support a cascade view of transfer. Cross-lingual lexical generalization depends on prior acquisition of several lower-level capabilities: the model must learn the shared grammar, learn how the minority language realizes that grammar, and learn the ontology-level type constraints before it can map unseen \LBm{} forms onto concepts observed through \LA{}. Failure at any earlier stage limits masked transfer. Additional experiments are shown in Appendix \ref{sec:appendix}.

%% file: sections/3.1.theory.tex
Our results suggest that transfer depends on whether the tokenizer leaves masked minority-language forms connected to observed forms through shared subword fragments. For a masked item $v \in \mathcal{V}^\dagger$, we define its \textit{tokenizer bridge} as

\[
    \beta_\pi(v)=\max_{u\in \mathcal{V}\setminus \mathcal{V}^\dagger}
    \mathrm{sim}\big(\pi(\rho_{\color{Bcolor} \mathbf{B}}(v)), \pi(\rho(u))\big),
\]

Where $\mathrm{sim}$ is normalized token overlap between tokens of the masked symbols and any other token seen during training. The average bridge strength $\hat{\beta}_\pi$ measures how isolated masked \LBm{} forms are under the tokenizer.

\begin{figure}[hbt!]
    \centering
    \includesvg[width=0.85\linewidth]{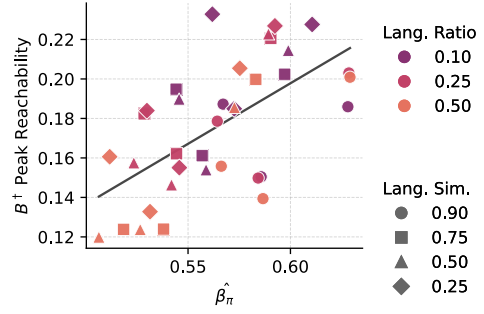}
    \caption{\textbf{Tokenizer bridges correlate with masked cross-lingual transfer.} We compare average tokenizer bridge strength $\hat{\beta}_\pi$ against peak Top-$K$ reachability in the masked minority-language condition \LBm{}. The positive correlation suggests that masked transfer improves when the tokenizer preserves subword paths between observed and unseen lexical realizations.}
    \label{fig:theory}
\end{figure}

This quantity explains the main empirical pattern. Smaller vocabularies often preserve reusable fragments, while larger vocabularies merge frequent strings into language-specific units, reducing bridge strength. As shown in Fig.~\ref{fig:theory}, $\hat{\beta}_\pi$ correlates strongly with peak masked reachability ($r=0.62$, $p<0.0001$), suggesting that masked transfer is enabled by subword paths between observed and unseen lexical realizations.

%% file: sections/5.conclusions.tex
We introduced an \textit{in-vitro} framework for studying cross-lingual generalization under controlled multilingual conditions. By procedurally generating two languages that share the same ontology, grammar, and compositional structure while differing in surface realization, we isolated factors that are normally entangled in natural corpora: lexical distance, language proportion, tokenizer training regime, and tokenizer vocabulary size. The masked minority-language condition further allowed us to test whether models can generalize to unseen lexical forms, rather than merely memorization.

Our results suggest that transfer depends on whether the tokenizer preserves reusable subword structure that connects observed and unseen forms across languages. Larger vocabularies can improve monolingual efficiency, but they may also collapse words into language-specific atoms, weakening the subword bridges needed for cross-lingual abstraction. Conversely, smaller vocabularies can support transfer by keeping lexical forms decomposable into fragments that remain shared across languages.

More broadly, our results support a staged view of multilingual learning. Models first acquire structural competence, such as grammaticality and type constraints, before they can reliably generalize lexical knowledge to unseen minority-language forms. These findings imply that multilingual language models should not be trained with tokenizer design treated as a preprocessing detail. The tokenizer defines the interface through which surface forms become available to the model, and therefore shapes which cross-lingual generalizations are reachable. 

%% file: sections/6.limitations.tex
Our framework makes several simplifying assumptions about the nature of language that limit its generalizability. First, we assume all languages conform to a recursive, context-sensitive grammar. This excludes languages such as (for example) Pirahã, which has been argued to lack recursive embedding, or languages commonly
referred to Khoisan, whose structural properties fall outside our grammatical assumptions \citep{everett2005cultural,khosian}. Similarly, Semitic languages such as Arabic and Hebrew employ non-concatenative morphology (root-and-pattern templatic systems \cite{mccarthy1990prosodic}) which our agglutinative word formation model cannot capture. We leave extension to these morphological types as future work.

Second, our lexical realization model assumes words are formed by concatenating prefixes, stems, and suffixes, and that letter probabilities follow a Zipfian distribution \cite{zipf1935}. This excludes fusional morphology, where morphemes merge and lose their boundaries, as well as tonal and phonological. We further abstract away from phonetics entirely, meaning that surface similarity is measured purely at the character level rather than through phonological categories or syllable structure. We instantiate our experiments with an SVO grammar, whereas natural languages exhibit several other word-order types, including SOV, VSO, and VOS. This is a limitation of the present configuration rather than the framework: other word orders can be supported by changing the grammar while keeping the ontology and evaluation protocol fixed.

Third, we assume that both languages share exactly the same concept ontology and vocabulary size. This rules out phenomena such as lexical gaps, where a concept expressible in one language has no direct equivalent in another, and false friends \cite{bentivogli2000looking,dinu-etal-2023-robocop}, where surface-similar forms carry different meanings across languages. Natural multilingual settings rarely satisfy these assumptions, and the degree to which our findings extend to such settings remains an open question.

Finally, language is inherently fluid and socially situated, and no finite generative grammar can fully capture its complexity. Our setup should therefore be understood as a controlled approximation designed to isolate specific transfer mechanisms, rather than a faithful model of natural language acquisition or multilingual competence.

%% file: sections/7.appendix.tex
\input{tables/examples}

\begin{figure}[hbt!]
    \centering
    \includesvg[width=1.0\linewidth]{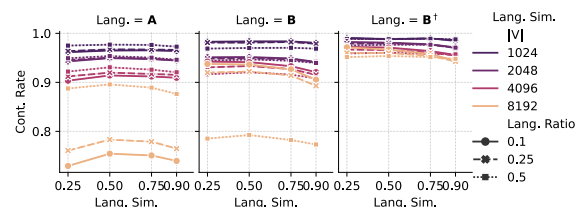}
    \caption{Tokenizer fragmentation continuation rate for \LA{}, \LB{}, and \LBm{}. The continuation rate measures how often words are represented by multi-token segmentations. Higher continuation rates indicate greater fragmentation. The masked condition follows the minority-language tokenization pattern, showing that the model must generalize through the same subword interface used for observed \LB{} forms.}
    \label{fig:fragmentation-cont}
\end{figure}

\subsection{Implementation details}
We keep the number of concepts fixed, corresponding to 100 entities, 10 classes to divide over, 460 descriptive properties, 40 descriptive values, 100 relative properties (see \cite{lubana2024percolation}). When forming symbolic sentences, we sampled concepts uniformly. Words are created using the following syllable patterns: "CVC", "CCV", "CVCC", "CV", "VC", "V". The letter probabilities follow a Zipfian distribution \cite{zipf1935} and are slightly different for \LA{} and \LB{}, according to the lexical distance $d$. The average word length is nine characters. 

The PCSG used to generate sentences is presented in \ref{fig:production-grammar}. Each symbolic vocabulary item is assigned a latent proto-stem and two surface realizations, one per language, obtained by applying language-specific cognate transformations and optional category-conditioned affixes; symbolic sequences are then realized by simple token substitution into the selected language (Algorithm \ref{alg:lexicon}. Here, $s$ denotes a latent proto-stem, $c(v)$ is the grammatical category of symbolic entry $v$, $\pi$ and $\sigma$ are an optional prefix and suffix, and $\Vert$ denotes string concatenation. The function $\textsc{DeriveCognate}(s,\ell)$ produces a language-specific variant of the shared stem $s$ for language $\ell$.

\input{algorithms/language-realisation}

Each configuration is trained across nine seeds: three seeds for the dataset creation and three seeds for the model training.  Model is a small transformer decoder \cite{vaswani2017attention} consisting of 4 layers, a model dimension of 256, using SwiGLU \cite{shazeer2020glu} and RMSNorm \cite{zhang2019root} layers. We used a batch size of 64, learning rate of 0.0001, adapted using a cosine decay with linear warmup of 256 steps. We used AdamW optimizer \cite{kingma2014adam} with $\beta_1 = 0.9$, $\beta_2 = 0.95$,   $\varepsilon = 10^{-10}$ and weight decay of 0.01. Context size was set at 256 tokens, enough such that sequences are not truncated. We train each model for 10000 steps.We train a total of 700 models. A training run takes around one hour. Model training was performed on a consumer NVIDIA RTX 3060 GPU, making our experiments replicable on consumer hardware. 

\input{algorithms/reachability}

\noindent \textbf{Top-K Reachability} Algorithm \ref{alg:reachable-topk-trie} checks whether any target sequence is reachable when decoding is restricted to tokens that are both valid trie continuations and present in the model's top-$k$ predictions at each step. By batching all active prefixes in the frontier, it evaluates many partial candidates with a single forward pass per decoding step.

\begin{figure}[t]
\centering
\begin{minipage}{0.95\linewidth}
\begin{tcolorbox}[
    colback=gray!6,
    colframe=gray!45,
    boxrule=0.4pt,
    arc=2pt,
    left=4pt,
    right=4pt,
    top=4pt,
    bottom=4pt,
    fontupper=\footnotesize\ttfamily,
    title={Production grammar},
    coltitle=black,
    colbacktitle=gray!15,
    fonttitle=\footnotesize\bfseries,
]
\begin{verbatim}
S     -> Ph NP VP EndOfSeq [1.0]
      | Ph NP VP SepSeq S  [0.0]

NP    -> subjectID         [0.8]
      | NP Conj NP         [0.2]

VP    -> descPreP descV    [0.4]
      | relV relPreP relNP [0.4]
      | VP Conj VP         [0.2]

relNP -> objectID          [0.7]
      | objectID Conj relNP[0.3]

Ph        -> '[P]'      [1.0]
subjectID -> 'subjectID'[1.0]
objectID  -> 'objectID' [1.0]
relV      -> 'relV'     [1.0]
descV     -> 'descV'    [1.0]
descPreP  -> 'descPreP' [1.0]
relPreP   -> 'relPreP'  [1.0]
Conj      -> 'conj'     [1.0]
SepSeq    -> '<sep>'    [1.0]
EndOfSeq  -> '<eos>'    [1.0]
\end{verbatim}
\end{tcolorbox}
\end{minipage}
\caption{Probabilistic grammar used to generate symbolic sequences. The grammar defines the abstract sentence structure shared by both languages before language-specific lexical realization. This shared symbolic layer allows us to isolate transfer through surface forms while holding syntax, type constraints, and ontology fixed.}
\label{fig:production-grammar}
\end{figure}

\subsection{Additional experiments.}

\begin{figure}[hbt!]
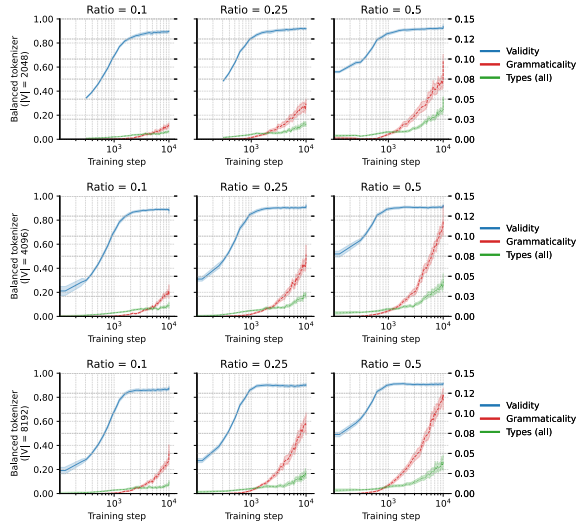

    \centering
    \includesvg[width=1.0\linewidth]{figures/results/condensed-grammar-B-masked-balanced-vocab2048.svg}
    \includesvg[width=1.0\linewidth]{figures/results/condensed-grammar-B-masked-balanced-vocab4096.svg}
    \includesvg[width=1.0\linewidth]{figures/results/condensed-grammar-B-masked-balanced-vocab8192.svg}
    \caption{Training dynamics for validity, grammaticality, and type-constraint satisfaction in the masked condition \LBm{} using a balanced tokenizer.}
    \label{fig:condensed-appendix}
\end{figure}

\begin{figure}[hbt!]
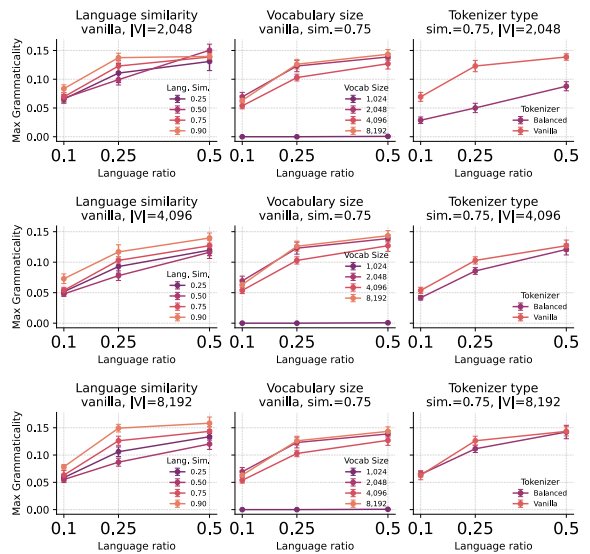

    \centering
    \includesvg[width=\linewidth]{figures/results/max-perf-max-grammaticality-vanilla-2048-0.25.svg}
    \includesvg[width=\linewidth]{figures/results/max-perf-max-grammaticality-vanilla-4096-0.25.svg}
    \includesvg[width=\linewidth]{figures/results/max-perf-max-grammaticality-vanilla-8192-0.25.svg}
    \caption{Grammaticality in the vanilla tokenizer setting across vocabulary sizes, language ratios, and lexical similarities. Grammaticality improves with additional minority-language data and is comparatively robust across lexical similarity conditions, indicating that structural well-formedness can be learned even when masked lexical transfer remains limited.}
    \label{fig:max-performance-grammaticality-vanilla}
\end{figure}

\begin{figure}[hbt!]
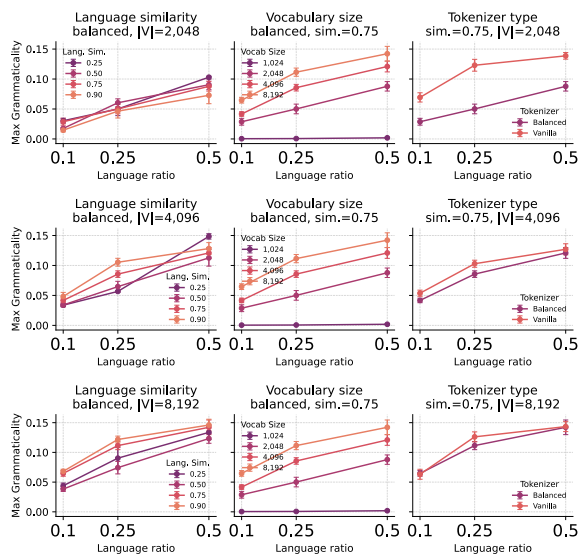

    \centering
    \includesvg[width=\linewidth]{figures/results/max-perf-max-grammaticality-balanced-2048-0.25.svg}
    \includesvg[width=\linewidth]{figures/results/max-perf-max-grammaticality-balanced-4096-0.25.svg}
    \includesvg[width=\linewidth]{figures/results/max-perf-max-grammaticality-balanced-8192-0.25.svg}
    \caption{Grammaticality in the balanced tokenizer setting across vocabulary sizes, language ratios, and lexical similarities. Balanced tokenizer training reduces some minority-language asymmetries, but its effect depends on vocabulary size. The results show that balanced tokenizer exposure does not automatically translate into uniformly better structural competence.}
    \label{fig:max-performance-grammaticality-balanced}
\end{figure}

\begin{figure}[hbt!]
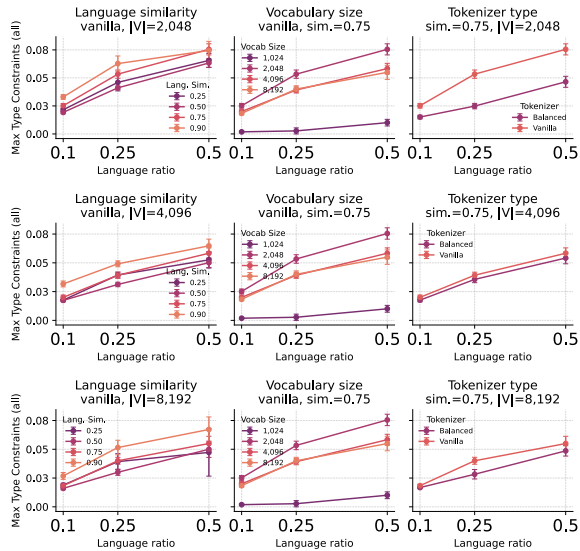

    \centering
    \includesvg[width=\linewidth]{figures/results/max-perf-max-type-constraints-all-vanilla-2048-0.25.svg}
    \includesvg[width=\linewidth]{figures/results/max-perf-max-type-constraints-all-vanilla-4096-0.25.svg}
    \includesvg[width=\linewidth]{figures/results/max-perf-max-type-constraints-all-vanilla-8192-0.25.svg}
    \caption{Type-constraint satisfaction in the vanilla tokenizer setting. Despite skewed tokenizer training, models acquire type-level competence across a range of conditions. The remaining gap between \LB{} and \LBm{} indicates that satisfying abstract constraints is easier than transferring those constraints to unseen target-language surface forms.}
    \label{fig:max-performance-types-vanilla}
\end{figure}

\begin{figure}[hbt!]
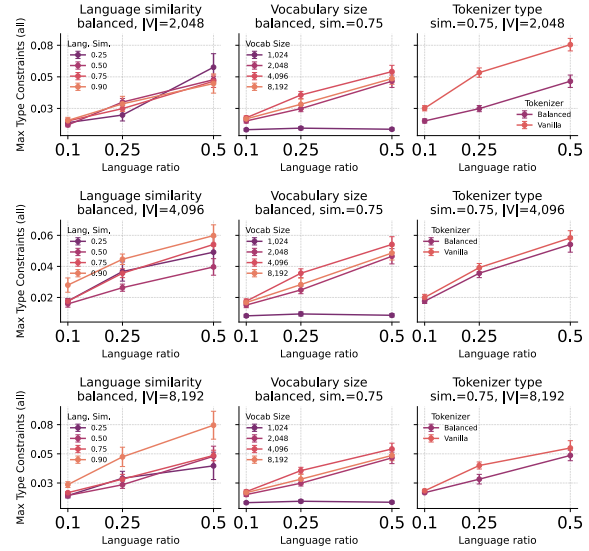

    \centering
    \includesvg[width=\linewidth]{figures/results/max-perf-max-type-constraints-all-balanced-2048-0.25.svg}
    \includesvg[width=\linewidth]{figures/results/max-perf-max-type-constraints-all-balanced-4096-0.25.svg}
    \includesvg[width=\linewidth]{figures/results/max-perf-max-type-constraints-all-balanced-8192-0.25.svg}
    \caption{Type-constraint satisfaction in the balanced tokenizer setting. Type-level competence improves with greater minority-language proportion and varies across vocabulary sizes. Compared with masked reachability, type satisfaction emerges more reliably, suggesting that semantic constraints are learned before they become usable for generating unseen minority-language lexical forms.}
    \label{fig:max-performance-types-balanced}
\end{figure}

\begin{figure}[hbt!]
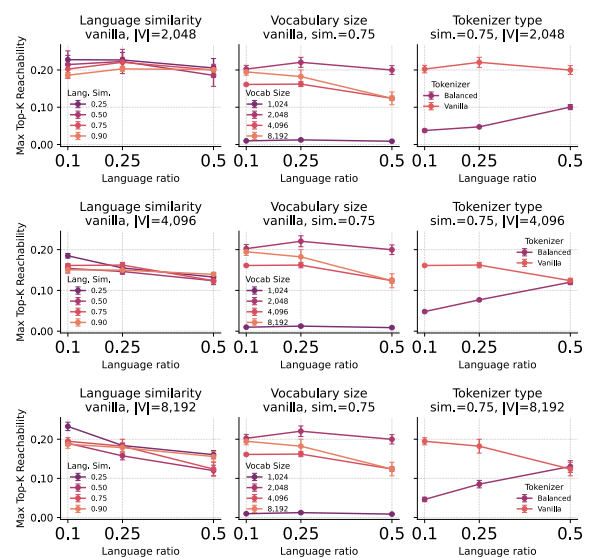

    \centering
    \includesvg[width=\linewidth]{figures/results/max-perf-max-top-k-reachability-vanilla-2048-0.25.svg}
    \includesvg[width=\linewidth]{figures/results/max-perf-max-top-k-reachability-vanilla-4096-0.25.svg}
    \includesvg[width=\linewidth]{figures/results/max-perf-max-top-k-reachability-vanilla-8192-0.25.svg}
    \caption{Masked Top-$K$ reachability in the vanilla tokenizer setting. Reachability is highest at smaller vocabulary sizes and generally decreases as vocabulary size increases. This supports the view that larger BPE vocabularies can harm cross-lingual transfer by replacing reusable subword fragments with language-specific lexical units.}
    \label{fig:max-performance-reachability-vanilla}
\end{figure}

\begin{figure}[hbt!]
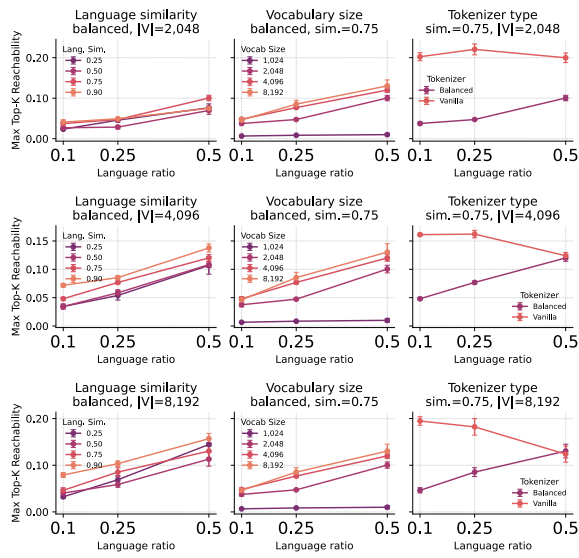

    \centering
    \includesvg[width=\linewidth]{figures/results/max-perf-max-top-k-reachability-balanced-2048-0.25.svg}
    \includesvg[width=\linewidth]{figures/results/max-perf-max-top-k-reachability-balanced-4096-0.25.svg}
    \includesvg[width=\linewidth]{figures/results/max-perf-max-top-k-reachability-balanced-8192-0.25.svg}
    \caption{Masked Top-$K$ reachability in the balanced tokenizer setting. Balanced tokenization can improve transfer when the minority-language proportion is small, but the benefit depends strongly on vocabulary size. Larger vocabularies again reduce reachability, showing that equal tokenizer exposure is not sufficient when the induced vocabulary fails to preserve shared subword structure.}
    \label{fig:max-performance-reachability-balanced}
\end{figure}

In Figure \ref{fig:condensed-appendix} we should training dynamics using a balanced tokenizer. Further in Figures \ref{fig:lang-capabilities-vanilla} and \ref{fig:lang-capabilities-balanced} we show emergence steps across languages for vanilla and balanced tokenizers, respectively. In Figures 
\ref{fig:max-performance-grammaticality-vanilla}, 
\ref{fig:max-performance-grammaticality-balanced}, 
\ref{fig:max-performance-types-vanilla}, 
\ref{fig:max-performance-types-balanced},
\ref{fig:max-performance-reachability-vanilla}, and
\ref{fig:max-performance-reachability-balanced}
we show the maximum performance reached across vanilla and balanced tokenizers.

\begin{figure}[hbt!]
    \centering
    \includesvg[width=1.0\linewidth]{figures/results/language-capabilities-vanilla-2048.svg}
    \includesvg[width=1.0\linewidth]{figures/results/language-capabilities-vanilla-4096.svg}
    \includesvg[width=1.0\linewidth]{figures/results/language-capabilities-vanilla-8192.svg}
    \caption{Average emergence time for grammaticality and type-constraint satisfaction across languages for vanilla tokenizer and $|V| \in \{2048, 4096, 8192\}$. Lower values indicate earlier emergence during training.}
    \label{fig:lang-capabilities-vanilla}
\end{figure}

\begin{figure}[hbt!]
    \centering
    \includesvg[width=1.0\linewidth]{figures/results/language-capabilities-balanced-2048.svg}
    \includesvg[width=1.0\linewidth]{figures/results/language-capabilities-balanced-4096.svg}
    \includesvg[width=1.0\linewidth]{figures/results/language-capabilities-balanced-8192.svg}
    \caption{Average emergence time for grammaticality and type-constraint satisfaction across languages for balanced tokenizer and $|V| \in \{2048, 4096, 8192\}$. Lower values indicate earlier emergence during training.}
    \label{fig:lang-capabilities-balanced}
\end{figure}

%% file: tables/examples.tex
\definecolor{tokAOnly}{HTML}{8BB9E5}
\definecolor{tokBOnly}{HTML}{ED979E}
\definecolor{tokSpecial}{HTML}{E9ECEF}
\definecolor{tokShared0}{HTML}{F4D35E}
\newcommand{\tokbox}[2]{%
  \begingroup
  \setlength{\fboxsep}{0.0pt}%
  \colorbox{#1}{\rule[-3pt]{0pt}{10pt}\texttt{\scriptsize #2}}%
  \endgroup
}
\begin{table*}[t]
\centering
\caption{Paired sentence realizations from the synthetic bilingual corpus chosen to emphasize visible overlap under the 2048 tokenizer. Each row comes from the same underlying symbolic sentence realized in \LA{} and \LB{}. Shared subtokens receive the same color in both languages. Legend: \tokbox{tokShared0}{shared} shared subtoken; \tokbox{tokAOnly}{A-only} \LA{}-only subtoken; \tokbox{tokBOnly}{B-only} \LB{}-only subtoken; \tokbox{tokSpecial}{<eos>} control token.}
\resizebox{\linewidth}{!}{\begin{tabular}{>{\raggedright\arraybackslash}p{0.49\linewidth}|>{\raggedright\arraybackslash}p{0.49\linewidth}}
\textbf{Language \LA{}} & \textbf{Language \LB{}} \\
\midrule
\tokbox{tokShared0}{Labzvae}␣ \tokbox{tokAOnly}{fekoc}␣ \tokbox{tokShared0}{x}\tokbox{tokShared0}{uc}\tokbox{tokShared0}{bo}\tokbox{tokShared0}{p}\tokbox{tokShared0}{z}\tokbox{tokAOnly}{h}\tokbox{tokAOnly}{by}\tokbox{tokShared0}{iw}␣ \tokbox{tokSpecial}{<eos>} & \tokbox{tokShared0}{Labzvae}␣ \tokbox{tokBOnly}{foc}␣ \tokbox{tokBOnly}{enux}\tokbox{tokShared0}{x}\tokbox{tokShared0}{uc}\tokbox{tokShared0}{bo}\tokbox{tokShared0}{p}\tokbox{tokShared0}{z}\tokbox{tokBOnly}{hi}\tokbox{tokBOnly}{v}\tokbox{tokShared0}{iw}\tokbox{tokBOnly}{big}␣ \tokbox{tokSpecial}{<eos>} \\
\addlinespace[0.2em]
\tokbox{tokShared0}{Nentmufkesi}␣ \tokbox{tokAOnly}{fekoc}␣ \tokbox{tokAOnly}{bbyadaec}␣ \tokbox{tokAOnly}{beqcq}\tokbox{tokShared0}{ic}\tokbox{tokShared0}{ijo}\tokbox{tokShared0}{bz}\tokbox{tokShared0}{z}\tokbox{tokShared0}{el}␣ \tokbox{tokAOnly}{lawj}␣ \tokbox{tokAOnly}{beqcmcb}␣ \tokbox{tokAOnly}{eose}␣ \tokbox{tokAOnly}{oe}␣ \tokbox{tokAOnly}{voyfeabwbadwi}␣ \tokbox{tokSpecial}{<eos>} & \tokbox{tokShared0}{Nentmufkesi}␣ \tokbox{tokBOnly}{foc}␣ \tokbox{tokBOnly}{guhibyodeil}␣ \tokbox{tokBOnly}{q}\tokbox{tokShared0}{ic}\tokbox{tokShared0}{ijo}\tokbox{tokShared0}{bz}\tokbox{tokShared0}{z}\tokbox{tokShared0}{el}␣ \tokbox{tokBOnly}{lasj}␣ \tokbox{tokBOnly}{mefb}␣ \tokbox{tokBOnly}{f}␣ \tokbox{tokBOnly}{ape}␣ \tokbox{tokBOnly}{vovyufefebwab}␣ \tokbox{tokSpecial}{<eos>} \\
\addlinespace[0.2em]
\tokbox{tokShared0}{Ditlethdehi}␣ \tokbox{tokAOnly}{fekoc}␣ \tokbox{tokShared0}{ik}\tokbox{tokShared0}{ad}\tokbox{tokAOnly}{ci}\tokbox{tokShared0}{z}\tokbox{tokShared0}{m}\tokbox{tokShared0}{oh}\tokbox{tokAOnly}{bbgi}␣ \tokbox{tokSpecial}{<eos>} & \tokbox{tokShared0}{Ditlethdehi}␣ \tokbox{tokBOnly}{foc}␣ \tokbox{tokShared0}{ik}\tokbox{tokShared0}{ad}\tokbox{tokBOnly}{ce}\tokbox{tokShared0}{z}\tokbox{tokShared0}{m}\tokbox{tokShared0}{oh}\tokbox{tokBOnly}{b}␣ \tokbox{tokSpecial}{<eos>} \\
\addlinespace[0.2em]
\tokbox{tokAOnly}{apyiefxnbac}␣ \tokbox{tokAOnly}{fekoc}␣ \tokbox{tokShared0}{d}\tokbox{tokShared0}{x}\tokbox{tokShared0}{ak}\tokbox{tokShared0}{of}\tokbox{tokShared0}{d}\tokbox{tokShared0}{ta}\tokbox{tokShared0}{n}\tokbox{tokAOnly}{bi}\tokbox{tokShared0}{t}␣ \tokbox{tokSpecial}{<eos>} & \tokbox{tokBOnly}{iynldwi}␣ \tokbox{tokBOnly}{foc}␣ \tokbox{tokBOnly}{enuxd}\tokbox{tokShared0}{x}\tokbox{tokShared0}{ak}\tokbox{tokShared0}{of}\tokbox{tokShared0}{d}\tokbox{tokShared0}{ta}\tokbox{tokShared0}{n}\tokbox{tokBOnly}{f}\tokbox{tokBOnly}{ga}\tokbox{tokShared0}{t}\tokbox{tokBOnly}{big}␣ \tokbox{tokSpecial}{<eos>} \\
\addlinespace[0.2em]
\tokbox{tokShared0}{Gecqulborzcpo}␣ \tokbox{tokAOnly}{fekoc}␣ \tokbox{tokAOnly}{occoabodbac}␣ \tokbox{tokAOnly}{ux}\tokbox{tokShared0}{va}\tokbox{tokShared0}{v}\tokbox{tokAOnly}{r}\tokbox{tokAOnly}{aw}\tokbox{tokShared0}{g}\tokbox{tokShared0}{jadom}␣ \tokbox{tokSpecial}{<eos>} & \tokbox{tokShared0}{Gecqulborzcpo}␣ \tokbox{tokBOnly}{foc}␣ \tokbox{tokBOnly}{accobabodbac}␣ \tokbox{tokShared0}{va}\tokbox{tokShared0}{v}\tokbox{tokBOnly}{re}\tokbox{tokBOnly}{w}\tokbox{tokShared0}{g}\tokbox{tokShared0}{jadom}␣ \tokbox{tokSpecial}{<eos>} \\
\addlinespace[0.2em]
\tokbox{tokShared0}{Segkligbmieq}␣ \tokbox{tokAOnly}{fekoc}␣ \tokbox{tokAOnly}{ogaarpcdwi}␣ \tokbox{tokShared0}{ej}\tokbox{tokShared0}{idd}\tokbox{tokShared0}{vu}\tokbox{tokShared0}{dom}␣ \tokbox{tokSpecial}{<eos>} & \tokbox{tokShared0}{Segkligbmieq}␣ \tokbox{tokBOnly}{foc}␣ \tokbox{tokBOnly}{gugiaralapcbac}␣ \tokbox{tokShared0}{ej}\tokbox{tokShared0}{idd}\tokbox{tokShared0}{vu}\tokbox{tokShared0}{dom}␣ \tokbox{tokSpecial}{<eos>} \\
\addlinespace[0.2em]
\tokbox{tokAOnly}{oibfisa}␣ \tokbox{tokAOnly}{hiikco}␣ \tokbox{tokAOnly}{fekoc}␣ \tokbox{tokAOnly}{occoabodbac}␣ \tokbox{tokAOnly}{bi}\tokbox{tokAOnly}{ir}\tokbox{tokShared0}{k}\tokbox{tokAOnly}{g}\tokbox{tokAOnly}{l}\tokbox{tokShared0}{ie}\tokbox{tokShared0}{vdom}␣ \tokbox{tokAOnly}{lawj}␣ \tokbox{tokAOnly}{fekoc}␣ \tokbox{tokAOnly}{bbyadaec}␣ \tokbox{tokShared0}{ie}\tokbox{tokShared0}{rf}\tokbox{tokShared0}{jo}\tokbox{tokShared0}{ob}\tokbox{tokAOnly}{bgi}␣ \tokbox{tokSpecial}{<eos>} & \tokbox{tokBOnly}{abfesadwi}␣ \tokbox{tokBOnly}{aihkco}␣ \tokbox{tokBOnly}{foc}␣ \tokbox{tokBOnly}{accobabodbac}␣ \tokbox{tokBOnly}{bab}\tokbox{tokShared0}{k}\tokbox{tokBOnly}{ga}\tokbox{tokBOnly}{et}\tokbox{tokBOnly}{vt}\tokbox{tokShared0}{ie}\tokbox{tokShared0}{vdom}␣ \tokbox{tokBOnly}{lasj}␣ \tokbox{tokBOnly}{foc}␣ \tokbox{tokBOnly}{guhibyodeil}␣ \tokbox{tokBOnly}{fiqeazaoe}\tokbox{tokShared0}{rf}\tokbox{tokShared0}{jo}\tokbox{tokShared0}{ob}␣ \tokbox{tokSpecial}{<eos>} \\
\addlinespace[0.2em]
\tokbox{tokShared0}{Gecqulborzcpo}␣ \tokbox{tokAOnly}{fekoc}␣ \tokbox{tokAOnly}{oabbabu}␣ \tokbox{tokAOnly}{beq}\tokbox{tokAOnly}{ced}\tokbox{tokShared0}{ue}\tokbox{tokShared0}{l}\tokbox{tokShared0}{ge}\tokbox{tokShared0}{gr}␣ \tokbox{tokSpecial}{<eos>} & \tokbox{tokShared0}{Gecqulborzcpo}␣ \tokbox{tokBOnly}{foc}␣ \tokbox{tokBOnly}{guabzaudwi}␣ \tokbox{tokBOnly}{enux}\tokbox{tokBOnly}{ed}\tokbox{tokShared0}{ue}\tokbox{tokShared0}{l}\tokbox{tokShared0}{ge}\tokbox{tokShared0}{gr}␣ \tokbox{tokSpecial}{<eos>} \\
\addlinespace[0.2em]
\tokbox{tokShared0}{Nentmufkesi}␣ \tokbox{tokAOnly}{fekoc}␣ \tokbox{tokAOnly}{bbyadaec}␣ \tokbox{tokAOnly}{beqcq}\tokbox{tokShared0}{ic}\tokbox{tokShared0}{ijo}\tokbox{tokShared0}{b}\tokbox{tokShared0}{z}\tokbox{tokShared0}{z}\tokbox{tokShared0}{el}␣ \tokbox{tokAOnly}{lawj}␣ \tokbox{tokAOnly}{beqcmcb}␣ \tokbox{tokAOnly}{eose}␣ \tokbox{tokAOnly}{oe}␣ \tokbox{tokAOnly}{voyfeabwbadwi}␣ \tokbox{tokSpecial}{<eos>} & \tokbox{tokShared0}{Nentmufkesi}␣ \tokbox{tokBOnly}{foc}␣ \tokbox{tokBOnly}{guhibyodeil}␣ \tokbox{tokBOnly}{q}\tokbox{tokShared0}{ic}\tokbox{tokShared0}{ijo}\tokbox{tokShared0}{b}\tokbox{tokShared0}{z}\tokbox{tokShared0}{z}\tokbox{tokShared0}{el}␣ \tokbox{tokBOnly}{lasj}␣ \tokbox{tokBOnly}{mefb}␣ \tokbox{tokBOnly}{f}␣ \tokbox{tokBOnly}{ape}␣ \tokbox{tokBOnly}{vovyufefebwab}␣ \tokbox{tokSpecial}{<eos>} \\
\addlinespace[0.2em]
\tokbox{tokAOnly}{oasifsgide}␣ \tokbox{tokAOnly}{oddwi}␣ \tokbox{tokAOnly}{lawj}␣ \tokbox{tokShared0}{Jeralogutc}␣ \tokbox{tokAOnly}{fekoc}␣ \tokbox{tokAOnly}{ruao}␣ \tokbox{tokAOnly}{if}\tokbox{tokAOnly}{l}\tokbox{tokAOnly}{oh}\tokbox{tokShared0}{ok}\tokbox{tokShared0}{z}\tokbox{tokShared0}{s}\tokbox{tokShared0}{id}\tokbox{tokShared0}{i}␣ \tokbox{tokSpecial}{<eos>} & \tokbox{tokBOnly}{apahbaqbegodwi}␣ \tokbox{tokBOnly}{obac}␣ \tokbox{tokBOnly}{lasj}␣ \tokbox{tokShared0}{Jeralogutc}␣ \tokbox{tokBOnly}{foc}␣ \tokbox{tokBOnly}{ae}␣ \tokbox{tokBOnly}{h}\tokbox{tokBOnly}{co}\tokbox{tokBOnly}{h}\tokbox{tokShared0}{ok}\tokbox{tokShared0}{z}\tokbox{tokShared0}{s}\tokbox{tokShared0}{id}\tokbox{tokBOnly}{ve}\tokbox{tokShared0}{i}␣ \tokbox{tokSpecial}{<eos>} \\
\addlinespace[0.2em]
\tokbox{tokShared0}{Labzvae}␣ \tokbox{tokAOnly}{fekoc}␣ \tokbox{tokAOnly}{x}\tokbox{tokShared0}{uc}\tokbox{tokShared0}{bo}\tokbox{tokShared0}{p}\tokbox{tokShared0}{zh}\tokbox{tokAOnly}{by}\tokbox{tokShared0}{iw}␣ \tokbox{tokSpecial}{<eos>} & \tokbox{tokShared0}{Labzvae}␣ \tokbox{tokBOnly}{foc}␣ \tokbox{tokBOnly}{enuxx}\tokbox{tokShared0}{uc}\tokbox{tokShared0}{bo}\tokbox{tokShared0}{p}\tokbox{tokShared0}{zh}\tokbox{tokBOnly}{iv}\tokbox{tokShared0}{iw}\tokbox{tokBOnly}{big}␣ \tokbox{tokSpecial}{<eos>} \\
\addlinespace[0.2em]
\tokbox{tokShared0}{Gecqulborzcpo}␣ \tokbox{tokAOnly}{lawj}␣ \tokbox{tokAOnly}{tciaqtabac}␣ \tokbox{tokAOnly}{apenjeca}␣ \tokbox{tokAOnly}{fekoc}␣ \tokbox{tokAOnly}{aphap}␣ \tokbox{tokAOnly}{uxv}\tokbox{tokAOnly}{hu}\tokbox{tokShared0}{ed}\tokbox{tokAOnly}{qu}\tokbox{tokShared0}{k}\tokbox{tokShared0}{ih}\tokbox{tokShared0}{le}␣ \tokbox{tokSpecial}{<eos>} & \tokbox{tokShared0}{Gecqulborzcpo}␣ \tokbox{tokBOnly}{lasj}␣ \tokbox{tokBOnly}{kadta}␣ \tokbox{tokBOnly}{aevnjeucdwi}␣ \tokbox{tokBOnly}{foc}␣ \tokbox{tokBOnly}{gunoap}␣ \tokbox{tokBOnly}{vh}\tokbox{tokShared0}{ed}\tokbox{tokBOnly}{uq}\tokbox{tokShared0}{k}\tokbox{tokShared0}{ih}\tokbox{tokBOnly}{ka}\tokbox{tokShared0}{le}␣ \tokbox{tokSpecial}{<eos>} \\
\addlinespace[0.2em]
\tokbox{tokShared0}{Gecqulborzcpo}␣ \tokbox{tokAOnly}{fekoc}␣ \tokbox{tokAOnly}{ux}\tokbox{tokShared0}{eh}\tokbox{tokAOnly}{ok}\tokbox{tokShared0}{mo}\tokbox{tokShared0}{ru}\tokbox{tokShared0}{h}\tokbox{tokAOnly}{hbgi}␣ \tokbox{tokSpecial}{<eos>} & \tokbox{tokShared0}{Gecqulborzcpo}␣ \tokbox{tokBOnly}{foc}␣ \tokbox{tokShared0}{eh}\tokbox{tokBOnly}{iv}\tokbox{tokShared0}{mo}\tokbox{tokShared0}{ru}\tokbox{tokShared0}{h}\tokbox{tokBOnly}{hbig}␣ \tokbox{tokSpecial}{<eos>} \\
\addlinespace[0.2em]
\tokbox{tokShared0}{Gecqulborzcpo}␣ \tokbox{tokAOnly}{fekoc}␣ \tokbox{tokAOnly}{oabbabu}␣ \tokbox{tokAOnly}{beq}\tokbox{tokAOnly}{ced}\tokbox{tokShared0}{ue}\tokbox{tokShared0}{l}\tokbox{tokShared0}{ge}\tokbox{tokShared0}{gr}␣ \tokbox{tokSpecial}{<eos>} & \tokbox{tokShared0}{Gecqulborzcpo}␣ \tokbox{tokBOnly}{foc}␣ \tokbox{tokBOnly}{guabzaudwi}␣ \tokbox{tokBOnly}{enux}\tokbox{tokBOnly}{ed}\tokbox{tokShared0}{ue}\tokbox{tokShared0}{l}\tokbox{tokShared0}{ge}\tokbox{tokShared0}{gr}␣ \tokbox{tokSpecial}{<eos>} \\
\addlinespace[0.2em]
\tokbox{tokShared0}{Nentmufkesi}␣ \tokbox{tokAOnly}{fekoc}␣ \tokbox{tokAOnly}{k}\tokbox{tokShared0}{ow}\tokbox{tokShared0}{c}\tokbox{tokShared0}{en}\tokbox{tokShared0}{rdom}␣ \tokbox{tokSpecial}{<eos>} & \tokbox{tokShared0}{Nentmufkesi}␣ \tokbox{tokBOnly}{foc}␣ \tokbox{tokBOnly}{fiqeazao}\tokbox{tokShared0}{ow}\tokbox{tokShared0}{c}\tokbox{tokShared0}{en}\tokbox{tokShared0}{rdom}␣ \tokbox{tokSpecial}{<eos>} \\
\addlinespace[0.2em]
\tokbox{tokShared0}{Segkligbmieq}␣ \tokbox{tokAOnly}{fekoc}␣ \tokbox{tokAOnly}{ib}\tokbox{tokShared0}{og}\tokbox{tokShared0}{b}\tokbox{tokAOnly}{ma}\tokbox{tokShared0}{vu}\tokbox{tokAOnly}{v}\tokbox{tokShared0}{ce}␣ \tokbox{tokSpecial}{<eos>} & \tokbox{tokShared0}{Segkligbmieq}␣ \tokbox{tokBOnly}{foc}␣ \tokbox{tokShared0}{b}\tokbox{tokShared0}{og}\tokbox{tokShared0}{b}\tokbox{tokBOnly}{pa}\tokbox{tokShared0}{vu}\tokbox{tokBOnly}{ta}\tokbox{tokBOnly}{s}\tokbox{tokShared0}{ce}\tokbox{tokBOnly}{big}␣ \tokbox{tokSpecial}{<eos>} \\
\bottomrule
\end{tabular}}
\end{table*}

%% file: algorithms/language-realisation.tex
\begin{algorithm}[t]
\caption{Bilingual lexical realization.}
\label{alg:lexicon}
\begin{algorithmic}[1]
\Require Symbolic vocabulary $\mathcal{V}$; symbolic sequence $X=(x_1,\dots,x_n)$; language set $\mathcal{L}=\{A,B\}$; category function $c:\mathcal{V}\to\mathcal{C}$
\Ensure Bilingual lexicon $M$ with entries $M[v,\ell]$ for $v\in\mathcal{V}$ and $\ell\in\mathcal{L}$

\Function{MakeWord}{$s, c, \ell$}
    \State $s^{(\ell)} \gets \textsc{DeriveCognate}(s,\ell)$
    \State $\pi \gets \textsc{SamplePrefix}(c,\ell)$
    \State $\sigma \gets \textsc{SampleSuffix}(c,\ell)$
    \State \Return $\pi \,\Vert\, s^{(\ell)} \,\Vert\, \sigma$
\EndFunction

\ForAll{$v \in \mathcal{V}$}
    \State sample latent stem $s$
    \ForAll{$\ell \in \mathcal{L}$}
        \State $M[v,\ell] \gets \textsc{MakeWord}(s, c(v), \ell)$
    \EndFor
\EndFor

\State \Return $M$
\end{algorithmic}
\end{algorithm}

%% file: algorithms/reachability.tex
\begin{algorithm}[t]
\caption{\textsc{ReachableTopK}}
\label{alg:reachable-topk-trie}
\begin{algorithmic}[1]
\Require model $f$, prompt $\mathbf{x}$, target sequences $\mathcal{S}$, top-$k$ value $k$, horizon $H$
\Ensure whether any target sequence is reachable

\State $T \gets \textsc{BuildTrie}(\mathcal{S})$
\State $\mathcal{F} \gets \{(r,\epsilon)\}$ \Comment{root node and empty prefix}

\For{$t = 1$ to $H$}
    \If{$\mathcal{F} = \emptyset$} \Return \textbf{false} \EndIf
    \State Batch all contexts $[\mathbf{x};\mathbf{p}]$ for $(n,\mathbf{p}) \in \mathcal{F}$
    \State Compute top-$k$ next-token sets $\mathcal{K}_{(n,\mathbf{p})}$
    \State $\mathcal{F}' \gets \emptyset$
    \ForAll{$(n,\mathbf{p}) \in \mathcal{F}$}
        \ForAll{$u \in \mathrm{Children}(n) \cap \mathcal{K}_{(n,\mathbf{p})}$}
            \State $n' \gets \mathrm{Next}(n,u)$
            \If{$n'$ is terminal} \Return \textbf{true} \EndIf
            \State add $(n', \mathbf{p}\|u)$ to $\mathcal{F}'$
        \EndFor
    \EndFor
    \State $\mathcal{F} \gets \textsc{Deduplicate}(\mathcal{F}')$
\EndFor
\State \Return \textbf{false}
\end{algorithmic}
\end{algorithm}